 \documentclass[pmlr]{jmlr}

 \RequirePackage{graphicx}
  \usepackage{booktabs}
  \usepackage{siunitx}

 \usepackage{longtable}
 \makeatletter
 \def\set@curr@file#1{\def\@curr@file{#1}} 
 \makeatother
 \usepackage[load-configurations=version-1]{siunitx} 
 
 \theorembodyfont{\upshape}
 \theoremheaderfont{\scshape}
 \theorempostheader{:}
 \theoremsep{\newline}

 \jmlrproceedings{PMLR}{Proceedings of Machine Learning Research}
 \jmlrvolume{340}
 \jmlryear{2026}
 \jmlrworkshop{Machine Learning for Healthcare}
 
\title[ANT]{Learning Prostate Anatomy at Test Time for Cancer Detection in Micro-Ultrasound}

\author{%
\begin{tabular*}{\textwidth}{@{}l@{\extracolsep{\fill}}r@{}}
\textbf{Obed Korshie Dzikunu}$^{1,2}$   & \mdseries\scshape obed@ece.ubc.ca \\
\textbf{Mohammad Mahdi Abootorabi}$^{1,2}$     & \mdseries\scshape mahdi.abootorabi@ece.ubc.ca \\
\textbf{Mohamed Harmanani}$^{2,3}$   & \mdseries\scshape mohamed.harmanani@queensu.ca \\
\textbf{Paul F. R. Wilson}$^{2,3}$   & \mdseries\scshape paul.wilson@queensu.ca \\
\textbf{Emma Willis}$^{2,3}$   & \mdseries\scshape emma.willis@queensu.ca \\
\textbf{Ferdinand Luger}$^{4}$   & \mdseries\scshape ferdinand.luger@ordensklinikum.at \\
\textbf{Adam Kinnaird}$^{5}$   & \mdseries\scshape ask@ualberta.ca \\
\textbf{Brian Wodlinger}$^{6}$   & \mdseries\scshape bwodlinger@exactimaging.com \\
\textbf{Parvin Mousavi}$^{2,3}$   & \mdseries\scshape mousavi@queensu.ca \\
\textbf{Purang Abolmaesumi}$^{1}$   & \mdseries\scshape purang@ece.ubc.ca \\
\end{tabular*}\\[0.5em]
\begin{minipage}{\textwidth}
\mdseries\footnotesize
$^{1}$ \textit{Department of Electrical and Computer Engineering, University of British Columbia, Vancouver, Canada}\\
$^{2}$ \textit{Vector Institute, Toronto, Canada}\\
$^{3}$ \textit{School of Computing, Queen's University, Kingston, Canada}\\
$^{4}$ \textit{Ordensklinikum Linz, Linz, Austria}\\
$^{5}$ \textit{Department of Surgery, University of Alberta, Edmonton, Canada}\\
$^{6}$ \textit{Exact Imaging, Markham, Canada}
\end{minipage}%
}

\begin{document}

\maketitle

\begin{abstract}
Domain shift across clinical centers using different imaging hardware or acquisition protocols remains a fundamental barrier to deploying deep learning models for prostate cancer (PCa) detection. Existing test-time adaptation (TTA) methods address distribution shift through entropy minimization or augmentation-based self-supervision, correcting for statistical differences in image appearance but ignoring the anatomical structure of the target domain. We propose \textbf{ANT}, a segmentation-guided TTA framework that adapts a pretrained cancer detection encoder to the target domain by solving an auxiliary prostate segmentation task at test time, supervised by pseudo-masks from a frozen pretrained segmentation network. By aligning encoder representations to prostate anatomy in the target domain, ANT corrects domain-specific feature drift while preserving cancer-discriminative structure. The model was trained on 693 patients imaged with an earlier-generation micro-ultrasound scanner in a multi-center clinical trial, and evaluated on
118 patients acquired with a newer-generation system across two centers in another clinical trial. Under a leave-one-center-out protocol with identical evaluation conditions across all methods, ANT improves mean AUC by 2.9\% and 3.6\% at the biopsy-core and patient levels, respectively, over no adaptation, outperforming TTA baselines. Code is available at: https://github.com/ObedDzik/ant.git.
\end{abstract}

\section{Introduction}
Prostate cancer (PCa) remains one of the most prevalent malignancies
among men worldwide \citep{rawla2019epidemiology}, and its diagnosis
relies on histopathological analysis of tissue obtained via
transrectal ultrasound (TRUS)-guided biopsy \citep{panzone2022transrectal}.
A fundamental challenge of conventional TRUS is its poor soft tissue
contrast at standard frequencies, which limits the ability of clinicians to
visually localize suspicious regions of interest. As a result,
systematic biopsy schemes sample multiple cores across the gland
in a spatially uniform pattern rather than targeting suspected
lesions, leading to frequent missed cancers and limiting diagnostic
sensitivity \citep{mate2023prostate}. Pre-procedural multi-parametric
MRI (mpMRI) partially addresses this by providing anatomical and
functional information for targeted biopsy guidance and represents
the current clinical gold standard for lesion localization
\citep{ahmed2017diagnostic}. However, its high cost, logistical
burden, and restricted availability limit its widespread adoption.

Micro-ultrasound addresses this gap by operating at higher
transducer frequencies (approximately $29$ MHz compared to
conventional $5 - 9$ MHz TRUS), producing spatial resolution
sufficient for the detection of cancer similar to that of mpMRI at a fraction of the cost, while
retaining the accessibility, portability, and real-time guidance
capabilities of ultrasound \citep{rohrbach2018high}. Specifically, clinical
studies have demonstrated that micro-ultrasound achieves
sensitivity for clinically significant PCa detection comparable
to mpMRI-guided biopsy \citep{eure2018comparison,klotz2020comparison,kinnaird2025microultrasonography}, 
positioning it as a viable and scalable alternative for targeted biopsy guidance. 

Deep learning models trained on micro-ultrasound have demonstrated the potential to automate cancer
identification \citep{gilany2023trusformer, wilson2024prostnfound, pensa2024deep, willis2026guideusgradeinformedunpaireddistillation} but 
models trained on data from a single
clinical center may suffer significant performance degradation when
deployed across different centers or scanner generations, a phenomenon known
as domain shift \citep{dataset_shift_2009, pooch2020can}. Test-time adaptation (TTA) offers a promising paradigm for addressing
this challenge without requiring labeled target domain data. By
adapting model parameters at inference time using a self-supervised
auxiliary signal, TTA methods reduce the gap between training and
deployment distributions \citep{sun2020test,wang2020tent}. Existing TTA methods
including TENT \citep{wang2020tent}, SAR \citep{niu2023towards}, EATA
\citep{niu2022efficient}, and CoTTA \citep{wang2022continual}, rely on entropy
minimization or augmentation consistency as self-supervised signals.
While effective for natural image distribution shifts, these signals do 
not exploit the structural and anatomical properties of medical images, properties which are clinically 
meaningful and tend to remain identifiable across domains even as image statistics vary substantially, as 
evidenced by the cross-domain success of generalist segmentation models \citep{MedSAM, wasserthal2023totalsegmentator}.

In prostate micro-ultrasound imaging, we hypothesize that the prostate gland remains a consistently 
identifiable anatomical target across patient populations, clinical centers, and scanner generations, 
even as low-level image statistics vary substantially due to differences in acquisition protocols 
and equipment. We further hypothesize that domain shift manifests as drift in the encoder's internal 
feature representations of this anatomy, corrupting the features that underpin downstream cancer 
detection. We therefore propose that supervising the encoder to segment the prostate at test time 
re-anchors these representations to this domain-invariant anatomical target, correcting feature drift 
without requiring target domain labels. This anatomically grounded signal provides explicit 
structural supervision at test time, in contrast to generic self-supervised objectives 
that derive their adaptation signal from the model's output distribution in the target domain, 
without leveraging domain-invariant properties of the input.

\begin{figure}[ht]
\begin{centering}
\includegraphics[width=\linewidth]{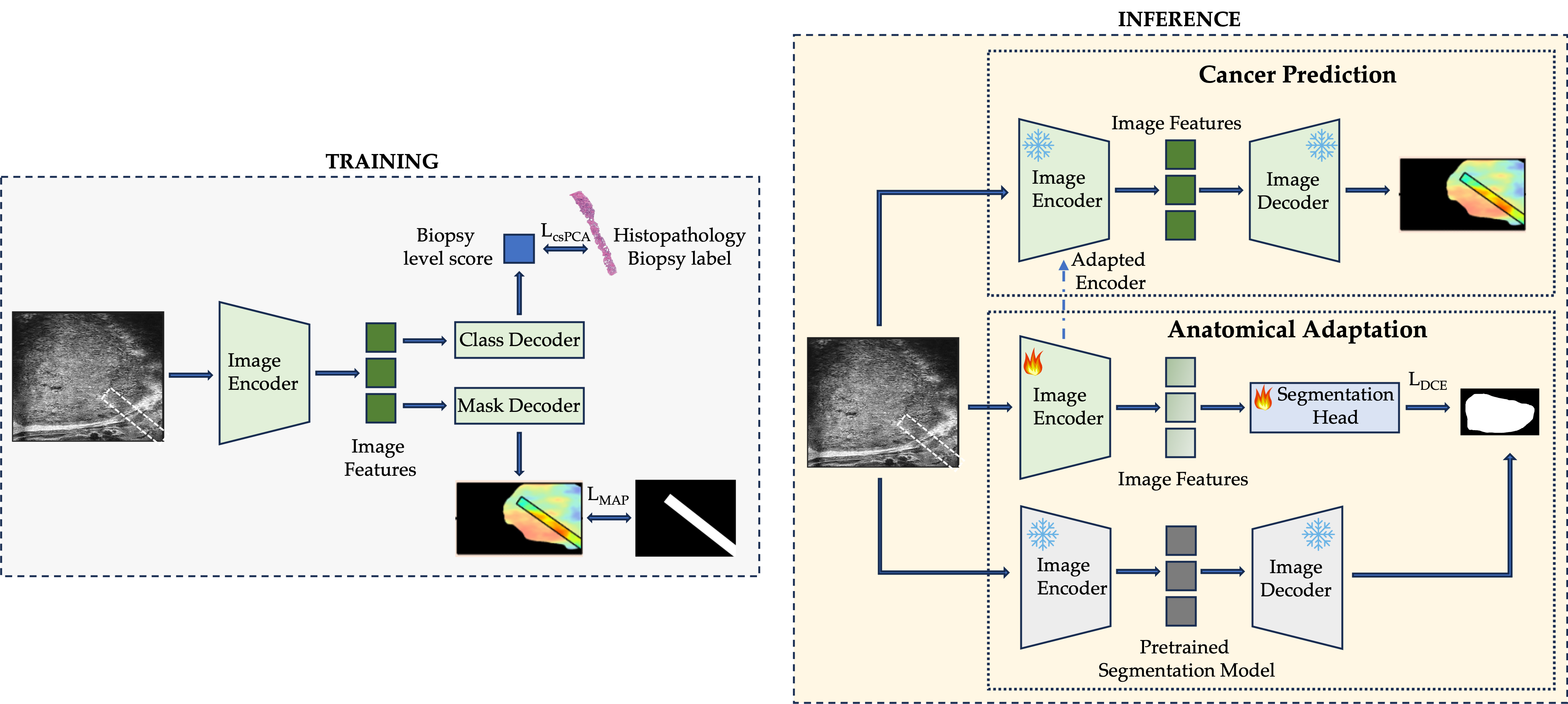}
\caption{Overview of ANT framework. At inference, the first $n$ layers of the trained cancer detection encoder is domain-adapted via an auxiliary segmentation task guided by pseudo-masks from a frozen segmentation network. The adapted encoder is then used for cancer
detection.
\label{fig:arch} 
 } 
 \end{centering}
\end{figure}

We therefore propose \textbf{ANT}, a TTA framework that adapts the encoder
of a pretrained cancer detection model by solving a prostate
segmentation auxiliary task at test time. A lightweight segmentation
head, guided by pseudo-masks from a frozen pretrained segmentation
network (MicroSegNet \citep{jiang2024microsegnet}), generates a supervision signal that drives
updates to the first $n$ layers of a Vision Transformer (ViT) backbone. 
The segmentation head accumulates adaptation across the test set, while  
the encoder is restored to its checkpoint state after each sample to prevent cross-sample leakage.
An overview of the proposed ANT framework during both the training and inference phases 
is illustrated in Figure \ref{fig:arch}.

Our contributions are as follows:
\begin{enumerate}
\item We propose a segmentation-guided TTA framework that leverages anatomical invariance to improve prostate cancer detection in micro-ultrasound. To the best of our knowledge, this is the first work to utilize organ segmentation as a test-time adaptation signal for cancer detection.
\item Across two multi-center clinical trial cohorts, we show that the proposed method consistently outperforms existing TTA approaches and improves upon a prior state-of-the-art prostate cancer detection model.
\item Through controlled ablation studies, we show that anatomical segmentation provides a robust adaptation signal for mitigating domain shift arising from variations in imaging data distributions.
\end{enumerate}

\subsection*{Generalizable Insights about Machine Learning in the Context of Healthcare}

This work offers the following insights relevant to machine learning
in healthcare beyond the specific application of prostate cancer
detection:
\begin{itemize}
\item {Anatomical invariance as a test time adaptation signal:}
In medical imaging, anatomical structures remain consistently identifiable across domains 
even when image statistics differ substantially, as evidenced by the success of generalist 
segmentation models that transfer across scanners, centers, and imaging modalities without 
domain-specific retraining \citep{MedSAM, wasserthal2023totalsegmentator}. The results of this work 
suggest that auxiliary tasks exploiting this identifiability, such as organ segmentation, 
can provide more targeted and stable adaptation signals than generic self-supervised objectives. 
While demonstrated here for prostate micro-ultrasound, we hypothesize that this principle may 
extend to other medical imaging applications where a domain-invariant anatomical supervision
target can be defined.

\item {Domain shift is encoded in early encoder layers:}
Our ablation results highlight that updating the first $n$ encoder
layers, which process low-level signal statistics,
outperforms updating middle or late layers. This suggests that cross-site and hardware-induced 
distribution shift may manifest as low-level feature perturbations, making the initial encoder 
layers an effective target for test-time adaptation.

\end{itemize}

\section{Related Work}
\subsection{Learning Strategies for Domain Shift}
A variety of approaches have been proposed to improve robustness under domain shift, 
broadly spanning domain adaptation, domain generalization, and 
meta-learning \citep{guan2021domain, yoon2024domain}. Domain adaptation 
methods assume access to labeled source data and unlabeled target data during 
training, and aim to reduce distribution mismatch via feature alignment, 
adversarial learning, or image translation \citep{ganin2016domain, kamnitsas2017unsupervised, chen2020unsupervised}. 
Domain generalization methods on the other hand learn representations that 
generalize to unseen domains without access to target data, using 
strategies such as invariant feature learning \citep{9543537} and 
data augmentation \citep{zhang2020generalizing, zhou2024mixstyle}. 
Meta-learning approaches further simulate domain shift during training by optimizing 
across source domain splits to improve robustness to unseen domains \citep{finn2017model, li2018learning}. 
While effective, these methods rely on training-time assumptions about domain variability and 
may not generalize to deployment settings where domain characteristics are unknown or evolving.

In prostate micro-ultrasound, domain shift may arise from differences in scanner generation, transducer frequency, 
and center-specific acquisition protocols, which may degrade model performance under deployment conditions. 
Existing approaches primarily address this through training-time design. 
For example, ProstNFound \citep{wilson2024prostnfound} integrates foundation models with 
ultrasound-specific features and clinical biomarkers to achieve strong multi-center 
performance on prostate cancer detection, while ProstNFound+ \citep{wilson2026prostnfound+} 
demonstrates improved generalisation to data acquired years later at a new clinical site. 
Although these methods achieve strong performance across the evaluated cohorts, they remain 
fixed after training and are therefore limited to the domain variability observed during development.
Similarly, while large-scale foundation models have shown robustness to cross-center variation in
prostate MRI \citep{li2024deep}, their behaviour under unseen micro-ultrasound 
acquisition conditions remain less well characterized. This motivates test-time 
adaptation methods, such as DEnEM \citep{gilany2024calibrated}, which adapt model 
predictions at inference time using only incoming data to better handle previously unseen domain shifts.

\subsection{Test-Time Adaptation}

TTA adjusts models at inference using unlabeled target-domain data, 
without access to source labels or retraining \citep{Liang_2024}. TENT \citep{wang2020tent} 
established the entropy minimization paradigm by updating batch normalization parameters at test time. 
Building on this, EATA \citep{niu2022efficient} introduces sample selection and Fisher regularization to 
prevent catastrophic forgetting, while SAR \citep{niu2023towards} improves stability under small 
batches via sharpness-aware minimization. CoTTA \citep{wang2022continual} extends TTA to continually 
shifting distributions through stochastic weight restoration and augmentation-averaged predictions, 
and ROID \citep{marsden2024universal} further addresses universal TTA through diversity-weighted loss 
and continual weight ensembling. MEMO \citep{zhang2022memo} takes a different approach, adapting 
each test sample independently via marginal entropy minimization over augmented views.

A recent large-scale benchmark of TTA methods in medical image
segmentation \citep{yu2025large} evaluated twenty
representative methods across seven imaging modalities under
unified protocols, finding that no single paradigm generalizes
across all conditions and that entropy-based methods degrade
significantly under large inter-center and inter-device shifts.
This finding is consistent with our observations in multi-center
micro-ultrasound and motivates the need for domain-specific
adaptation signals beyond generic entropy objectives.

More recently, task-specific TTA methods have been proposed for
medical imaging. GraTa \citep{chen2025gradient} improves adaptation by
aligning pseudo-task gradients with the primary task gradient
direction using dynamic learning rates. A3-TTA \citep{wu2025a3} applies anchor-guided
pseudo-label generation with boundary-aware entropy minimization
for prostate segmentation under domain shift. DEnEM \citep{gilany2024calibrated}
addresses domain shift in micro-ultrasound prostate cancer
detection through neural network ensembles for calibrated marginal
entropy minimization. ReservoirTTA \citep{vray2025reservoirtta}
maintains domain-specialised model reservoirs with online style
clustering for recurring distribution shifts.

Despite this progress, existing TTA methods rely on entropy
minimization, augmentation consistency, or gradient alignment as
adaptation signals, all of which operate on the global feature
statistics or output logits without exploiting domain-invariant
structural properties of the input. We address this
gap by using prostate segmentation as a structured, anatomically
grounded TTA signal that directly targets domain-specific feature drift in the encoder, 
providing a principled alternative to
entropy-based adaptation for clinical deployment under
multi-center domain shift.

 \section{Methods}
 \label{sec:method}
 
 \subsection{Problem Formulation}
 
 Let $f_{\mathrm{enc}}: \mathcal{X} \rightarrow \mathcal{Z}$ denote a
 pretrained encoder mapping micro-ultrasound images to feature
 representations, and let $g: \mathcal{Z} \rightarrow \mathcal{Y}$
 denote the detection head mapping features to cancer probability
 heatmaps. The model $h = g \circ f_{\mathrm{enc}}$ is trained on source
 domain data $\mathcal{D}_{\mathrm{src}} = \{(x_i, y_i)\}_{i=1}^{N}$
 from a clinical trial cohort noted as Center A.
 
 At test time, the model encounters target domain data
 $\mathcal{D}_{\mathrm{tgt}} = \{x_j\}_{j=1}^{M}$ from unseen centers in a different cohort
 with no available labels. The domain gap
 $\mathcal{G} = d(\mathcal{D}_{\mathrm{src}}, \mathcal{D}_{\mathrm{tgt}})$
 degrades detection performance. Our goal is to adapt $f_{\mathrm{enc}}$ to
 $\mathcal{D}_{\mathrm{tgt}}$ using only unlabeled test samples and an
 anatomical auxiliary task.
 
 \subsection{Anatomical Alignment via Segmentation TTA}
 \paragraph{Pseudo-label generation.}
 We use a pretrained frozen MicroSegNet \citep{jiang2024microsegnet}
 $s_{\phi}: \mathcal{X} \rightarrow [0,1]^{H \times W}$ to generate
 prostate pseudo-masks for each test image:
 \begin{equation}
   \hat{m}_j = \sigma\!\left(s_{\phi}(x_j)\right),
 \end{equation}
 where $\sigma$ denotes the sigmoid function. The MicroSegNet
 parameters $\phi$ are frozen throughout TTA.
 
 \paragraph{Segmentation head.}
 A lightweight segmentation head
 $p_{\psi}: \mathcal{Z} \rightarrow \mathbb{R}^{H' \times W'}$ is
 attached to the encoder output:
 \begin{equation}
   \hat{s}_j = p_{\psi}\!\left(f_{\mathrm{enc}}(x_j)\right),
 \end{equation}
 where $\hat{s}_j$ is upsampled to match $\hat{m}_j$ via bilinear
 interpolation. The head consists of two convolutional layers with
 GroupNorm and a final $1{\times}1$ projection to a single channel
 binary output. The head is randomly initialised and is not
 reset between test samples, allowing it to accumulate an increasingly
 refined understanding of target domain anatomy across the test set.
 
 \paragraph{TTA objective.}
 The TTA loss combines Dice loss and binary cross-entropy:
 \begin{equation}
   \mathcal{L}_{\mathrm{TTA}} = \mathcal{L}_{\mathrm{Dice}}(\hat{s}_j, \hat{m}_j)
   + \lambda \mathcal{L}_{\mathrm{BCE}}(\hat{s}_j, \hat{m}_j),
   \label{eq:loss}
 \end{equation}
 where $\lambda$ balances the two terms. Dice loss encourages global
 shape agreement between predicted and pseudo segmentation masks, while
 binary cross-entropy enforces pixel-level alignment.
 
 \subsection{Encoder Update Strategy}
 \label{sec:strategy}
 Rather than updating all encoder parameters, we selectively update
 all parameters of the first $n$ transformer blocks of the ViT
 backbone:
 \begin{equation}
   \Theta_{\mathrm{update}} = \left\{\theta_l \;:\;
   l \in \mathcal{I}_{\mathrm{f}}\right\},
   \label{eq:update}
 \end{equation}
 where $\theta_l$ denotes all parameters of block $l$ and
 $\mathcal{I}_{\mathrm{f}}$ denotes the indices of the first $n$
 blocks.
 
    \begin{algorithm2e}[t]
        \caption{\textsc{ANT}: Segmentation-Guided Test-Time Adaptation}
        \label{alg:anatta}
        \DontPrintSemicolon
        \SetAlgoLined
        \SetKwInput{KwIn}{Input}
        \SetKwInput{KwOut}{Output}
        \SetKwInput{KwState}{Persistent}
        
        \KwIn{Image $x_j$, checkpoint $W^*$, frozen segmentor $s_\phi$,
              steps $\tau$, lr $\eta$}
        \KwState{Seg head $\psi$, optimizer $\mathcal{O}$}
        \KwOut{Cancer score $\hat{y}_j$}
        
        \BlankLine
        $\hat{m}_j \leftarrow \sigma\!\left(s_\phi(x_j)\right)$
        \tcp*{pseudo-mask}
        
        $W_{\mathrm{enc}} \leftarrow W^*$
        \tcp*{initialise from checkpoint}
        
        \BlankLine
        \For(\tcp*[f]{adapt first $n$ blocks}){$t \leftarrow 1$ \KwTo $\tau$}{
            $\hat{s}_j \leftarrow \mathrm{Up}\!\left(p_\psi\!\left(f_{W_{\mathrm{enc}}}(x_j)\right)\right)$\;
            $\mathcal{L} \leftarrow \mathcal{L}_{\mathrm{Dice}}(\hat{s}_j, \hat{m}_j) + \lambda\,\mathcal{L}_{\mathrm{BCE}}(\hat{s}_j, \hat{m}_j)$\;
            $W_{\mathrm{enc}},\, \psi \leftarrow \mathcal{O}\!\left(\nabla_{\Theta, \psi}\,\mathcal{L}\right)$\;
        }
        
        \BlankLine
        $\hat{y}_j \leftarrow g\!\left(f_{W_{\mathrm{enc}}}(x_j)\right)$
        \tcp*{predict with adapted encoder}
        
        $W_{\mathrm{enc}} \leftarrow W^*$
        \tcp*{reset encoder, $\psi$ and $\mathcal{O}$ persist}
        
        \Return $\hat{y}_j$
    \end{algorithm2e}

\subsection{PCa Detection}
\subsubsection{Detection Model Architecture}

The cancer detection model consists of three components. The image
encoder is a pretrained DINOv3 ViT-L/16 backbone
\citep{simeoni2025dinov3} with $24$ transformer blocks, each
equipped with LayerNorm layers. Intermediate hidden states are
extracted at interaction indices $\{4,11, 17, 23\}$ and processed
by a UNETR-style hierarchical decoder \citep{hatamizadeh2022unetr}
comprising skip-connection encoder and decoder blocks that
progressively upsample features to produce a spatial cancer
probability heatmap.

The heatmap features are subsequently passed to a two-way
transformer \citep{kirillov2023segment} that attends jointly over
image embeddings and learnable query tokens, enabling bidirectional
cross-attention between spatial features and task-specific
representations. The output is processed by two task-specific
decoders: a mask decoder that produces a dense cancer probability
heatmap over the biopsy region, and a class decoder that produces
an image-level classification score for clinically significant
PCa. Cancer predictions at the biopsy core level are obtained by
averaging heatmap values along the biopsy needle trajectory.

During TTA, only the first $n$ transformer blocks of the DINOv3 encoder are updated via the segmentation auxiliary task. The UNETR
decoder, two-way transformer, mask decoder, and class decoder
remain frozen throughout adaptation.

\subsubsection{Training Objective}
\label{sec:training_objective}
The detection model is trained using the multi-task objective proposed by \citet{wilson2026prostnfound+}, combining image-level classification supervision with weak spatial constraints on the predicted heatmap. Image-level supervision is applied to the csPCa probability via a binary cross-entropy loss with the pathology label. In parallel, spatial supervision is derived from core-level involvement labels by aggregating the predicted heatmap within the biopsy needle region and comparing it to the corresponding involvement score using binary cross-entropy. The final objective is the sum of these two terms, jointly encouraging accurate core-level classification while promoting spatially coherent heatmap responses that reflect tumour burden within the sampled region.

Let $y_{\mathrm{cls}} \in \{0,1\}$ denote the core-level pathology label and $\hat{y}_{\mathrm{cls}} \in [0,1]$ the predicted probability. The model outputs a spatial heatmap $P \in [0,1]^{H \times W}$, which is aggregated over the biopsy region defined by a binary mask $M \in \{0,1\}^{H \times W}$. Let $\mathcal{N} = \{(i,j)\,|\,M_{i,j}=1\}$ denote the set of valid pixels within the needle region. The predicted cancer involvement is computed as a masked expectation:

\begin{equation}
\hat{I} = \frac{1}{|\mathcal{N}|} \sum_{(i,j)\in\mathcal{N}} P_{i,j}.
\end{equation}

The training objective combines core-level and region-level supervision:
\begin{equation}
\mathcal{L} = \mathcal{L}_{\mathrm{csPCa}} + \mathcal{L}_{\mathrm{map}},
\end{equation}
where
\begin{equation}
\mathcal{L}_{\mathrm{csPCa}} = \mathrm{BCE}(y_{\mathrm{cls}}, \hat{y}_{\mathrm{cls}}), \quad
\mathcal{L}_{\mathrm{map}} = \mathrm{BCE}(I, \hat{I}),
\end{equation}
and $I \in [0,1]$ denotes the ground-truth cancer involvement score.

\section{Study Cohort}

\subsection{Cohort Selection}
Data were derived from clinical trials with predefined inclusion
and exclusion criteria. Enrolled patients were indicated for prostate biopsy due to elevated PSA and/or abnormal digital rectal examination, were aged $\geq$18 years, and had no prior prostate biopsy or history of genitourinary cancer. Additional criteria included no contraindications to biopsy or mpMRI, and no recent mpMRI for prostate cancer evaluation. Exclusion criteria mirrored these conditions.

\subsection{Data Acquisition and Study Design}
The training cohort consisted of retrospective data collected between 2013 and 2016 from a multi-center clinical trial (NCT02079025) involving five institutions, collectively called Center A (source). Patients underwent untargeted systematic prostate biopsy using an earlier version of the ExactVu $\mu$US system (Exact Imaging, Markham, Canada), with typically $10-12$ cores acquired per subject \citep{rohrbach2018high}.

The test cohort comprised prospective data collected between 2021 and 2024 (NCT05220501) from two clinical sites using a newer version of the ExactVu $\mu$US system. In this cohort, patients underwent both systematic and targeted biopsies. Targeting was performed using either micro-ultrasound ($\mu$US) with PRI-MUS scoring or pre-procedural mpMRI with PI-RADS scoring, depending on the study arm. MRI–$\mu$US fusion was conducted using the ExactVu FusionVu system \citep{kinnaird2025microultrasonography}.

Biopsy locations were assigned PRI-MUS-based risk scores across both arms. In the MRI-guided arm, targeted cores inherited the PI-RADS score of the corresponding lesion; if no lesion was identified, a prostate-level PI-RADS score (1 or 2) from the radiology report was assigned to all cores from that patient. Summary statistics for both cohorts are provided in Table \ref{tab:cohort}.

\subsection{Data Extraction}

Sagittal B-mode micro-ultrasound images were acquired at $28$ mm
depth and $46.06$ mm width during ultrasound-guided biopsy.
A 30-second cineloop was recorded per core, capturing needle
alignment and firing. The needle trace was defined by annotating
the tip position and insertion angle at the moment of firing,
from which a corresponding path mask was generated. The final
frame immediately prior to needle entry, co-registered with the
needle trace, was selected as the representative image for each
biopsy core. This frame captures the pre-fire tissue state
without needle artefact and is the standard input for
learning-based analysis in this setting
\citep{wilson2026prostnfound+}.

Histopathological analysis of each biopsy specimen provided the
cancer diagnosis, percentage involvement (approximate fraction
of tumour length within the core), and ISUP Grade Group (GG $0-5$).
Clinically significant PCa (csPCa) was defined as GG$\geq$3;
cores with GG $1-2$ were categorised as insignificant PCa and
GG $0$ as benign. Labels were assigned to the full needle trace
region, introducing a minor spatial approximation for partially
involved cores. Patient-level labels reflected the most severe
core finding for that patient.

\subsection{Feature Choices}

Raw B-mode images were resized from their native resolution of
$1372 \times 833$ pixels to $512 \times 512$ pixels using
bilinear interpolation and normalised to the range $[0, 1]$.
All predictions were made soley from imaging features with no clinical metadata provided as input.

The resized images were passed to the DINOv3 ViT-L/16 encoder,
which divides each image into non-overlapping patches of
$16 \times 16$ pixels, yielding a sequence of
$32 \times 32 = 1024$ patch tokens each of dimension 1024.
These patch tokens are processed by the transformer blocks to
produce contextualized spatial feature representations, which
are extracted at interaction indices $\{4, 11, 17, 23\}$ and
passed to the UNETR decoder for heatmap generation. Cancer predictions at the biopsy core level are obtained by averaging
heatmap activations along the needle trace mask $\mathcal{N}$
as described in Section \ref{sec:training_objective}.

 \begin{table}[!t]
   \centering
   \caption{Cohort statistics per clinical center. Centers differ in
     scanner generation and acquisition protocol, representing
     realistic deployment-time domain shift.}
   \begin{tabular}{lccc}
     \toprule
     \textbf{Center} & \textbf{Patients} & \textbf{Cores} &
     \textbf{Cancer \%} \\
     \midrule
     A (source) & $693$ & $5479$ & $16\%$ \\
     B           & $76$ & $1118$ & $34.6\%$ \\
     C           & $42$ & $569$ & $27.0\%$ \\
     \bottomrule
   \end{tabular}
   \label{tab:cohort}
 \end{table}
 
\section{Results}
\label{sec:results}
\subsection{Evaluation Approach and Study Design}
 
The detection model is trained exclusively on Center A and evaluated under a leave-one-center-out (LOCO) protocol over target centers B and C, with one serving as the validation set for checkpoint selection and the other as the test set. Performance is reported as biopsy-core-level area under the curve value (AUC). At no point are target labels used to supervise adaptation; all methods operate exclusively on unlabeled test images. All baselines are reported using their published hyperparameter configurations. However, for EATA\citep{niu2022efficient} and SAR\citep{niu2023towards}, the entropy margin was rescaled from the ImageNet-calibrated value to account for binary classification following the same proportional logic as the original papers. TENT\citep{wang2020tent}, CoTTA\citep{wang2022continual}, ROID\citep{marsden2024universal}, and MEMO\citep{zhang2022memo} required no such rescaling and are reported directly under their published defaults.
 
\subsection{Comparison Against TTA Baselines}
 
Table \ref{tab:main_results} reports AUC on held-out test centers under the LOCO protocol. ANT achieves the highest mean core-level AUC and patient-level AUC for clinically significant cancer of $84.4\%$ and $82.1\%$, respectively, outperforming the no-adaptation baseline by $2.9\%$ and $3.6\%$ at the core and patient level, and surpassing all competing TTA methods. Notably, ANT achieves particularly strong performance on Center C, reaching $87.8\%$ core-level AUC and $84.3\%$ patient-level AUC and a sensitivity of $94.4\%$ and $83.1\%$  at specificity values of $60\%$ and $80\%$, respectively, at the core level (Table \ref{tab:sensitivity_results}).

Existing TTA methods show mixed and inconsistent results relative to the no-adaptation baseline. All baseline methods tend to improve on one center while degrading on the other, resulting in mean AUCs that are comparable to or below the no-adaptation baseline across most metrics. For instance, TENT achieves $80.1\%$ core-level csPCa AUC on Center B but drops to $81.9\%$ on Center C, and SAR shows a similar pattern with $80.9\%$ on Center B falling to $82.5\%$ on Center C. Moreover, ANT achieves statistically significant improvement over SAR, the second-best baseline, on core-level csPCa AUC at Center C. We attribute the superiority of ANT to the anatomical alignment signal, which provides a structured and domain-invariant supervision target that generic self-supervised signals cannot replicate.

\begin{table}[!t]
\centering
\caption{AUC (95\% bootstrap CI) for all methods across both evaluation 
centres (B and C). Bold indicates the best performance per column.}
\label{tab:main_results}
\resizebox{\textwidth}{!}{%
\begin{tabular}{lcccccc}
\toprule

& \multicolumn{2}{c}{\textbf{Patient-Level csPCa AUC}} 
& \multicolumn{2}{c}{\textbf{Core-Level csPCa AUC}} 
& \multicolumn{2}{c}{\textbf{Core-Level PCa AUC}} \\
\cmidrule(lr){2-3} \cmidrule(lr){4-5} \cmidrule(lr){6-7}

\textbf{Method}

& \textbf{B} & \textbf{C}
& \textbf{B} & \textbf{C}
& \textbf{B} & \textbf{C} \\
\midrule
No adaptation
    & $77.6 (72.9-82.0)$ & $79.4 (66.8-89.9)$
    & $78.5 (73.6-83.3)$ & $84.5 (74.0-93.1)$
    & $71.5 (68.2-74.6)$ & \textbf{73.8 (68.5 -- 79.3)} \\
MEMO \citep{zhang2022memo}
    & $78.3 (73.8-82.5)$ & $83.5 (73.0-91.7)$
    & $77.7 (73.1-82.5)$ & $83.4 (71.8-92.9)$
    & $73.3 (70.2-76.3)$ & $70.6 (65.3-76.3)$ \\
TENT \citep{wang2020tent}
    & $76.0 (71.6-80.2)$ & $75.0 (60.8-87.0)$
    & $80.1 (76.0-84.3)$ & $81.9 (70.3-92.6)$
    & $74.5 (71.5-77.3)$ & $68.6 (62.7-74.4)$ \\
EATA \citep{niu2022efficient}
    & $73.5 (68.7-77.2)$ & $69.4 (65.1-72.8)$
    & $79.8 (75.3-82.4)$ & $80.5 (77.3-84.2)$
    & \textbf{74.9 (71.0 -- 77.2)} & $71.9 (69.8-74.5)$ \\
CoTTA \citep{wang2022continual}
    & $77.9 (73.2-82.2)$ & $77.9 (64.6-89.7)$
    & $78.5 (73.7-83.3)$ & $84.1 (73.8-92.7)$
    & $71.4 (68.2-74.5)$ & $72.9 (67.4-78.4)$ \\
ROID \citep{marsden2024universal} 
    & $78.2 (73.8-82.2)$ & $76.8 (64.1-87.6)$
    & $79.9 (75.4-84.3)$ & $84.4 (73.0-92.9)$
    & $74.3 (71.3-77.3)$ & $71.0 (65.3-76.6)$ \\
SAR \citep{niu2023towards}
    & $79.5 (75.2-83.7)$ & $79.2 (68.1-89.5)$
    & $80.9 (76.4-85.5)$ & $82.5 (72.5-91.1)$
    & $73.3 (70.3-76.4)$ & $70.5 (64.9-75.8)$ \\
\midrule
\textbf{ANT (ours)}
    & \textbf{79.8 (75.6 -- 83.7)} & \textbf{84.3 (73.4 -- 92.8)}
    & \textbf{81.0 (76.8 -- 85.0)} & \textbf{$^*$87.8 (76.4 -- 96.0)}
    & $73.0 (69.9-76.2)$         & $72.3 (66.7-77.7)$ \\
\bottomrule
\end{tabular}%
}
{\footnotesize $^*$Core-Level csPCa AUC at Centre C: ANT vs SAR, DeLong test $p = 0.044$.}
\end{table}

\begin{table}[h]
\centering
\caption{Sensitivity at fixed specificity (95\% bootstrap CI) for all 
methods across both evaluation centres (B and C) at the core level for 
clinically significant prostate cancer (csPCa). Bold indicates the best 
performance per column.}
\label{tab:sensitivity_results}
\resizebox{\textwidth}{!}{%
\begin{tabular}{lcccc}
\toprule

& \multicolumn{2}{c}{\textbf{Sensitivity @ 60\% Specificity (\%)}}
& \multicolumn{2}{c}{\textbf{Sensitivity @ 80\% Specificity (\%)}} \\

\cmidrule(lr){2-3} \cmidrule(lr){4-5}

{\textbf{Method}}

& \textbf{B} & \textbf{C}
& \textbf{B} & \textbf{C} \\
\midrule
No adaptation
    & $81.0 (73.5-88.2)$ & $85.3 (43.7-100.0)$
    & $67.1 (56.1-76.5)$ & $70.1 (43.7-94.1)$ \\
MEMO \citep{zhang2022memo}
    & $82.5 (75.3-89.4)$ & $85.2 (64.7-100.0)$
    & $61.7 (51.3-72.1)$ & $74.5 (50.0-95.6)$ \\
TENT \citep{wang2020tent}
    & $80.4 (72.7-87.5)$ & $83.1 (47.1-100.0)$
    & $66.4 (57.8-75.4)$ & $71.1 (47.1-91.7)$ \\
EATA \citep{niu2022efficient}
    & $80.2 (71.3-86.4)$ & $89.5 (49.2-100.0)$
    & $65.5 (57.3-73.2)$ & $63.2 (44.6-88.5)$ \\
CoTTA \citep{wang2022continual}
    & $80.0 (72.0-87.3)$ & $83.4 (53.3-100.0)$
    & $65.8 (55.8-75.5)$ & $75.2 (53.3-94.4)$ \\
ROID \citep{marsden2024universal}
    & \textbf{83.4 (76.7 -- 90.2)} & $94.0 (55.6-100.0)$
    & $68.7 (60.2-77.2)$ & $77.4 (55.6-95.9)$ \\
SAR \citep{niu2023towards}
    & $81.9 (74.5-88.8)$ & $88.5 (43.5-100.0)$
    & \textbf{70.9 (62.5 -- 79.5)} & $72.2 (43.5-94.1)$ \\
\midrule
\textbf{ANT (ours)}
    & \textbf{83.4 (76.8 -- 90.1)} & \textbf{94.4 (64.3 -- 100.0)}
    & $69.7 (60.5-78.5)$ & \textbf{83.1 (64.3 -- 100.0)} \\
\bottomrule
\end{tabular}%
}
\end{table}

\subsection{Sensitivity to Pseudo-Mask Quality}
To assess ANT's sensitivity to pseudo-mask quality, ground truth prostate segmentation masks were obtained for 32 randomly sampled cases spanning both evaluation centres. MicroSegNet Dice scores were computed against these masks and ranged from 0.36 to 0.99 (median 0.83). Cases were stratified into tertiles based on Dice score: low ($\leq$0.72), mid 
$(0.72-0.90)$, and high ($>$0.90). Per-patient AUC was computed for PCa and csPCa separately for ANT and the unadapted baseline.

Table \ref{tab:dice_sensitivity} summarizes the results. Despite 
substantially imperfect segmentation in the low Dice tertile (median 
Dice = 0.60, range $0.36-0.71$), ANT consistently improves over the 
unadapted baseline on the clinically primary metric, csPCa AUC, across 
all three tertiles (3/3, 5/5, and 4/6 cases respectively; 12/14 overall). 
Of the 16 cases (similar performance on 3 cases) where ANT underperforms on PCa AUC, ANT recovers superiority on csPCa AUC in 5/6 of those cases with available csPCa data. Thus, ANT's adaptation signal remains beneficial for clinically significant disease detection across the full range of observed mask quality.

\begin{table}[!t]
\centering
\caption{Per-patient AUC for ANT and the unadapted baseline stratified 
by MicroSegNet pseudo-mask Dice score tertile. Values reported as median 
(range). csPCa AUC is available only for patients with at least one 
positive and one negative core for clinically significant cancer.}
\label{tab:dice_sensitivity}
\resizebox{\textwidth}{!}{%
\begin{tabular}{l cc ccc ccc}
\toprule
&&& \multicolumn{3}{c}{\textbf{PCa AUC (\%)}}
& \multicolumn{3}{c}{\textbf{csPCa AUC (\%)}} \\
\cmidrule(lr){4-6} \cmidrule(lr){7-9}

\textbf{Dice Tertile} & \textbf{$n$} & \textbf{Dice Median (Range)}  

&\textbf{ANT} & \textbf{Baseline} & \textbf{ANT $>$ Base}
&\textbf{ANT} & \textbf{Baseline} & \textbf{ANT $>$ Base} \\
\midrule
Low $(\leq0.72)$  & 11 & $0.60 (0.36-0.71)$ & 
    $80.0 (29.6-100.0)$ & $78.6 (23.3-100.0)$ & 4/11 & 
    $100.0 (78.1-100.0)$ & $80.0 (65.6-82.7)$ & 3/3 \\
Mid $(0.72-0.90)$ & 10 & $0.83 (0.72-0.90)$ & 
    $68.3 (6.2-100.0)$  & $76.2 (18.2-100.0)$ & 3/10 & 
    $63.6 (50.0-95.8)$  & $54.5 (36.7-70.8)$  & 5/5 \\
High $(>0.90)$   & 11 & $0.95 (0.91-0.99)$ & 
    $57.1 (21.4-94.4)$  & $44.0 (21.4-96.3)$  & 6/11 & 
    $71.7 (40.0-100.0)$ & $65.7 (23.1-100.0)$ & 4/6 \\
\midrule
\textbf{Overall} & \textbf{32} & \textbf{$0.83 (0.36-0.99)$} & 
    - & - & \textbf{13/32} & 
    - & - & \textbf{12/14} \\
\bottomrule
\end{tabular}%
}
\end{table}

\subsection{Generalization to State-of-the-Art Architecture}
To demonstrate that ANT is model-independent, we apply it to
ProstNFound+ (PNF+) \citep{wilson2026prostnfound+}, the current
state-of-the-art model for prostate cancer detection in
micro-ultrasound. PNF+ uses a MedSAM \citep{MedSAM} image encoder with
adapter-based fine-tuning and a prompt-driven class decoder.
We apply ANT to the PNF+ encoder following the same protocol
as in Section \ref{sec:strategy}, updating the first $n$ encoder blocks
using segmentation pseudo-masks from the frozen MicroSegNet, without retuning any hyperparameters.

Tables \ref{tab:prostnfound} and \ref{tab:prostnfound_sensitivity} report performance of PNF+ with and
without ANT on the held-out test centers. ANT improves PNF+ performance by a mean AUC of at least $3\%$ across
test centers, confirming that anatomical alignment via segmentation
provides a consistent domain adaptation benefit regardless of the
underlying detection architecture. Notably, this improvement is
achieved without any architecture-specific tuning.

\begin{table}[!t]
  \centering
  \caption{ANT applied to ProstNFound+ (PNF+), the current
    state-of-the-art model for micro-ultrasound prostate cancer
    detection. Results (AUC, $\%$) demonstrate model-independent generalization
    of the proposed TTA approach. Best results in bold.}
  \small
  \resizebox{\textwidth}{!}{
  \begin{tabular}{l cc cc cc}
    \toprule
    & \multicolumn{2}{c}{\textbf{Patient-Level csPCa AUC}}
    & \multicolumn{2}{c}{\textbf{Core-Level csPCa AUC}}
    & \multicolumn{2}{c}{\textbf{Core-Level PCa AUC}} \\
    
    \cmidrule(lr){2-3} \cmidrule(lr){4-5} \cmidrule(lr){6-7}
    
    \textbf{Method}
    & \textbf{Center B} & \textbf{Center C}
    & \textbf{Center B} & \textbf{Center C}
    & \textbf{Center B} & \textbf{Center C} \\
    \midrule
    PNF+ \citep{wilson2026prostnfound+}
    & $71.2 (66.8-75.6)$ & $75.6 (64.4-85.3)$
    & $72.4 (68.2-76.5)$ & $75.4 (62.8-87.2)$
    & $69.8 (66.6-73.1)$ & $63.8 (58.1-69.4)$ \\
    \textbf{PNF+ + ANT (ours)}
    & \textbf{72.8 (68.3 -- 77.4)} & \textbf{80.5 (69.7 -- 88.8)}
    & \textbf{73.9 (69.8 -- 78.0)} & \textbf{78.2 (66.5 -- 87.1)}
    & \textbf{$^*$73.1 (70.4 -- 76.3)} & \textbf{65.2 (59.3 -- 70.5)} \\
    \bottomrule
  \end{tabular}
  \label{tab:prostnfound}
  }
{\footnotesize $^*$Core-Level PCa AUC at Centre B: ANT vs PNF+, DeLong test $p < 0.001$.}
\end{table}

\begin{table}[!t]
\centering
\caption{Sensitivity at fixed specificity (95\% bootstrap CI) for the 
unadapted ProstNFound baseline (PNF) and ANT across both evaluation 
centres (B and C) at the core level for clinically significant prostate 
cancer (csPCa).}
\label{tab:prostnfound_sensitivity}
\resizebox{\textwidth}{!}{%
\begin{tabular}{lcccc}
\toprule

& \multicolumn{2}{c}{\textbf{Sensitivity @ 60\% Specificity (\%)}}
& \multicolumn{2}{c}{\textbf{Sensitivity @ 80\% Specificity (\%)}} \\

\cmidrule(lr){2-3} \cmidrule(lr){4-5}
{\textbf{Method}}

& \textbf{B} & \textbf{C}
& \textbf{B} & \textbf{C} \\
\midrule
PNF+ \citep{wilson2026prostnfound+}
    & $76.4 (68.6-84.3)$ & $72.3 (50.0-92.9)$
    & $44.5 (33.3-56.0)$ & \textbf{58.4 (30.4 -- 87.5)} \\
\textbf{PNF+ + ANT (ours)}
    & \textbf{78.9 (70.6 -- 86.4)} & \textbf{85.5 (65.0 -- 100.0)}
    & \textbf{46.0 (36.2 -- 56.7)} & $56.0 (30.8-77.8)$ \\
\bottomrule
\end{tabular}%
}
\end{table}

\begin{figure}[ht]
\begin{centering}
\includegraphics[width=\linewidth]{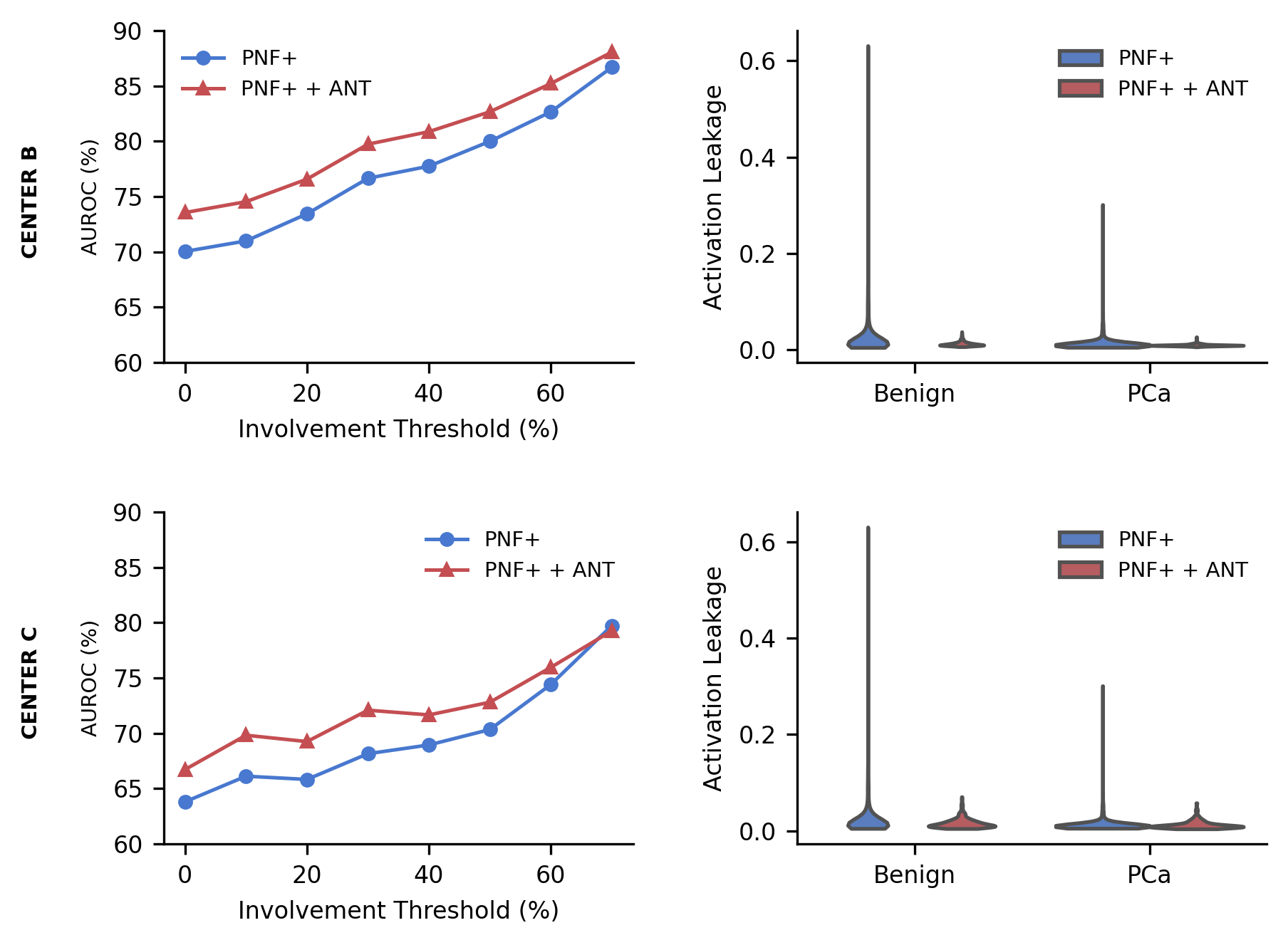}
\caption{Left: AUROC for clinically significant cancer prediction evaluated across increasing involvement 
thresholds (cores with true involvement $\geq t$ or benign). Right: Distribution of activation leakage, 
defined as the ratio between average prostate heatmap activation and average needle-region activation, 
stratified by cancer status (benign vs PCa) and method, PNF+ (no TTA) vs. PNF+ + ANT. Lower values indicate 
better spatial specificity, i.e., reduced off-target activation outside the biopsy needle region.
\label{fig:activation} 
 } 
 \end{centering}
\end{figure}

\begin{figure}[!t]
\begin{centering}
\includegraphics[width=12cm]{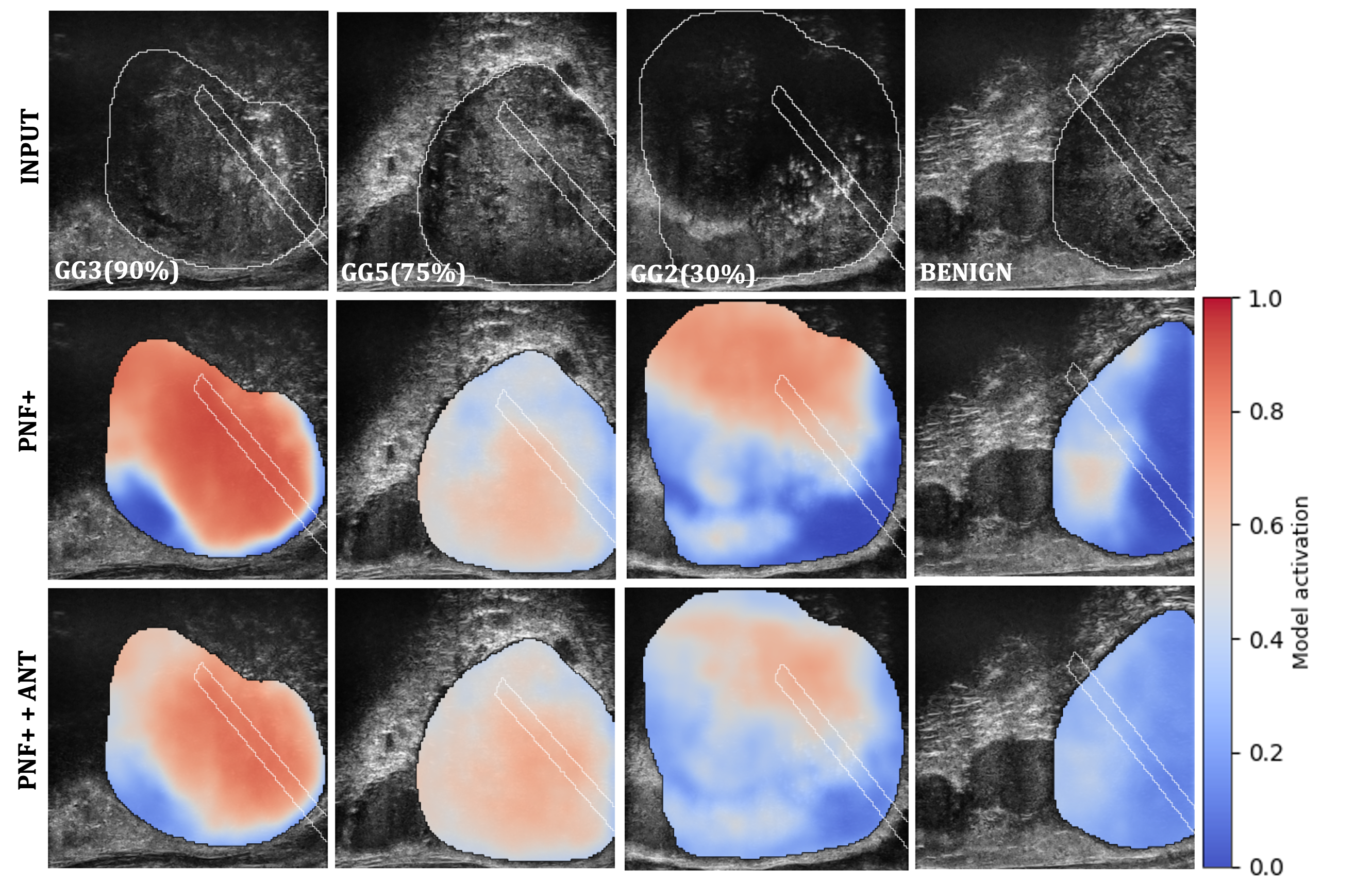}
\caption{Representative heatmaps of model activations without TTA (PNF+) and with TTA (PNF+ + ANT) on the same biopsy cores.
Activations represent predicted involvement, with higher values indicating greater likelihood of clinically significant cancer.
Each example is annotated with grade group (GG) and corresponding involvement percentage.
\label{fig:residual} 
 } 
 \end{centering}
\end{figure}

Figure \ref{fig:activation} shows the AUROC for clinically significant cancer detection across increasing involvement thresholds and the distribution of activation leakage stratified by tissue type and model, for both Center B and Center C. Both models demonstrate improved discriminative performance as the involvement threshold increases, with AUROC rising from approximately $70-74\%$ at threshold $0\%$ to $87-88\%$ at $70\%$ for Center B, and from $63-67\%$ to $79-80\%$ for Center C. PNF+ + ANT consistently outperforms PNF+ across all thresholds and both centers, suggesting that the proposed test time adaptation method improves the model's ability to distinguish clinically significant cancer from benign cores, particularly at lower involvement thresholds where the task is hardest. Regarding spatial specificity, activation leakage, defined as the ratio of average prostate heatmap activation to average needle-region activation, is markedly reduced by ANT across both centers and tissue types. While PNF+ exhibits a pronounced upper tail in the leakage distribution, reflecting frequent off-target activations outside the biopsy needle region, PNF+ + ANT concentrates nearly all of its mass near zero with no comparable tail, indicating substantially tighter spatial confinement of model activations. This improvement in anatomical specificity is consistent across both benign and PCa cores and both centers, demonstrating that ANT not only boosts classification performance but also produces more spatially precise heatmap activations. Qualitative comparisons of these heatmap activations, demonstrating tighter spatial confinement with the proposed method, are shown in Figure \ref{fig:residual}.

\subsection{Ablation Studies}
\label{sec:ablation}
 
\subsubsection{Encoder Update Strategy}
 
Table \ref{tab:ablation_layers} reports the effect of varying which
encoder blocks and parameters are updated during TTA. Our method
updates all parameters of the first $n$ blocks. We additionally
compare against restricting updates to normalization parameters only
or bias terms only, and against updating middle or late blocks.
 
\begin{table}[!t]
  \centering
  \caption{Ablation on encoder update strategy and auxiliary task.
    Mean AUC ($\%$) across held-out test centers. Best in bold.}
  \footnotesize
  \begin{tabular}{ll ccc}
    \toprule
    \textbf{Update target} & \textbf{Position}
    & \textbf{\shortstack{Patient-Level\\(csPCa)}}
    & \textbf{\shortstack{Core-Level\\(csPCa)}}
    & \textbf{\shortstack{Core-Level\\(PCa)}} \\
    \midrule
    \multicolumn{5}{l}{\textit{Layer position and parameter type}} \\
    \midrule
    Full blocks   & First $n$   & \textbf{82.1} & \textbf{84.4} & 72.7 \\
    Full blocks   & Middle $n$  & $79.8$ & $81.9$ & \textbf{73.6} \\
    Full blocks   & Last $n$    & $77.2$ & $80.8$ & $72.7$ \\
    Full blocks   & All blocks  & $74.9$ & $79.6$ & $70.2$ \\
    Norm only     & First $n$   & $76.4$ & $80.9$ & $72.8$ \\
    Norm + bias   & First $n$   & $80.4$ & $82.2$ & $73.0$ \\
    Bias only     & First $n$   & $78.8$ & $80.6$ & $71.8$ \\
    \midrule
    \multicolumn{5}{l}{\textit{Auxiliary task signal (first $n$ blocks, full update)}} \\
    \midrule
    Reconstruction & First $n$  & $76.5$ & $80.7$ & $72.5$ \\
    \textbf{Segmentation (ours)} & \textbf{First $n$}
      & \textbf{82.1} & \textbf{84.4} & \textbf{72.7} \\
    \bottomrule
  \end{tabular}
  \label{tab:ablation_layers}
\end{table}
 
Updating the first $n$ encoder blocks yields the strongest performance across all evaluation metrics, 
outperforming updates applied to middle, late, or all blocks. Adapting the last $n$ blocks or the full network,
consistently degrades performance, suggesting that these blocks capture 
task-specific representations that are sensitive to perturbation from the auxiliary objective. 
In contrast, early blocks appear to provide a more effective locus for adaptation.

Restricting updates to subsets of parameters within the first $n$
blocks (e.g., normalization layers or bias terms) provides modest improvements but does 
not match the performance of full block updates. This indicates that partial 
parameter adaptation is insufficient, and that modifying the full set of 
parameters within early blocks is important for effective feature realignment. 
Finally, the results highlights a difference in performance based on the choice of auxiliary task, with 
anatomical alignment yielding stronger robustness to domain shift.
 
\subsubsection{Number of TTA Steps}
 
Table \ref{tab:ablation_steps} reports performance as a function of
the number of inner optimization steps $\tau$. Performance varies non-monotonically
with the number of inner optimization steps $\tau$
Increasing $\tau$ from $2$ to $10$ leads to consistent improvements across all metrics, 
with peak performance achieved at $\tau=10$. However, further increasing the number 
of steps to $20$ results in a noticeable decline, particularly at the core level. 
This suggests that a moderate number of adaptation steps is sufficient to achieve 
effective domain alignment, while excessive optimization may lead to overfitting or 
degradation of task-relevant features.

 \begin{table}[!t]
    \centering
    \caption{Ablation on number of TTA inner steps. Mean AUC ($\%$) across
     held-out test centers. Best in bold.}
    \footnotesize
    \begin{tabular}{cccc}
      \toprule
      \textbf{Inner steps $\tau$}
      & \textbf{\shortstack{Patient-Level\\(csPCa)}}
      & \textbf{\shortstack{Core-Level\\(csPCa)}}
      & \textbf{\shortstack{Core-Level\\(PCa)}} \\
      \midrule
        0(w/o ANT) & $78.5$ & $81.5$ & \textbf{$72.7$} \\
        2    & $76.0$ & $81.1$ & $72.6$ \\
        5    & $78.9$ & $83.4$ & $72.5$ \\
        10   & \textbf{82.1} & \textbf{84.4} & \textbf{72.7} \\
        20   & $78.2$ & $78.0$ & $72.0$ \\
      \bottomrule
    \end{tabular}
    \label{tab:ablation_steps}
  \end{table}
 
 \section{Discussion}
This work shows that anatomically grounded test-time adaptation (TTA) via prostate segmentation improves both diagnostic performance and spatial specificity in micro-ultrasound prostate cancer detection under multi-center domain shift. Across unseen clinical centers and scanner generations, the proposed ANT method consistently outperforms both the non-adapted baseline and a strong state-of-the-art prostate cancer detection model (PNF+), while also improving anatomical alignment of model activations.

A consistent pattern emerges across all baseline TTA methods: performance improves at Center B (prevalence: $34.6\%$) but degrades at Center C (prevalence: $27.0\%$). This asymmetry is not straightforwardly explained by class prior shift alone as Center C has a prevalence closer to the source ($16\%$) than Center B, yet shows greater degradation. Notably, ROID \citep{marsden2024universal}, which explicitly incorporates prior correction to address class distribution shift, exhibits the same pattern, further suggesting that prevalence shift is not the primary driver. This observation may thus point to a more fundamental limitation: adaptation signals that do not anchor to domain-invariant structure may remain susceptible to center-specific confounds, limiting their reliability under heterogeneous deployment conditions. The proposed ANT framework addresses this limitation by using prostate segmentation as a structured anatomical prior during test-time adaptation, which unlike entropy-based approaches, provides pixel-level spatial supervision that directly constrains adaptation to the organ of interest.

This interpretation is further supported by the reduction in activation leakage, defined as the ratio of average prostate heatmap activation to needle-region activation. As shown in Figure \ref{fig:activation}, while PNF+ exhibits a pronounced high-leakage tail indicating frequent off-target activations, PNF+ + ANT concentrates its distribution near zero across both benign and cancerous cores and both test centers. This suggests that ANT improves spatial specificity without sacrificing disease sensitivity as excessive off-target activation may reflect spurious correlations and undermine clinician trust. Additionally, the generalization of ANT to PNF+ without architecture-specific tuning further demonstrates that anatomical test-time adaptation is backbone-agnostic.

A central hypothesis of this work is that the prostate gland remains a consistently identifiable anatomical target across patient populations, clinical centers, and scanner generations, even as low-level image statistics vary substantially. The results of this study provide empirical support for this hypothesis: MicroSegNet generates pseudo-masks of sufficient quality across both test centers and the full range of patient populations encountered, as demonstrated by the pseudo-mask sensitivity analysis (Table \ref{tab:dice_sensitivity}), where ANT consistently improves over the unadapted baseline on csPCa AUC across all Dice tertiles including the low quality tertile (median Dice $= 0.60$). This suggests that the prostate gland is identifiable enough across the domain shifts encountered in this study to serve as a reliable anatomical supervision target, providing a more stable adaptation signal than entropy-based objectives.

A potential concern with ANT's adaptation strategy is whether updating early encoder blocks while freezing downstream layers introduces a feature distribution mismatch that limits adaptation effectiveness. We attribute the consistent improvements observed across both test centers to two factors. First, gradients act on early blocks whose outputs propagate through many subsequent frozen layers before reaching the decoder, attenuating any representational mismatch at the decoder input. Second, the decoder was trained on augmented source domain images, providing implicit robustness to small feature distribution shifts introduced by test-time encoder updates. Together, these properties suggest that early-block adaptation via a segmentation objective produces feature shifts that are both gradual and within the decoder's operating range, rather than introducing disruptive distributional discontinuities.

\paragraph{Limitations.}
ANT relies on a pretrained prostate segmentation model to generate pseudo-labels. While the pseudo-mask sensitivity analysis (Table \ref{tab:dice_sensitivity}) demonstrates that adaptation remains beneficial across the range of segmentation quality observed in this study, these results are limited to the domain shifts encountered between the source and two target centers. Whether the prostate gland remains sufficiently identifiable, and ANT therefore effective under more severe domain shifts, such as substantially different scanner generations, imaging protocols, or patient populations, remains an open question that future work should address.

 
 \bibliography{tta_paper}
 
 \newpage
 \appendix
\section{Implementation Details}

\subsection{Cancer Detection Model}

\subsubsection{Training Configuration}

All experiments were run using PyTorch $2.1$ on a single NVIDIA
L40S GPU ($48$ GB). The detection model was trained for $35$ epochs
with a batch size of $4$ using the AdamW optimizer ($\beta_1=0.9$, $\beta_2=0.999$, $\epsilon=10^{-8}$, weight decay $=0.01$),
with a fixed random seed of 30. The encoder (DINOv3 ViT-L/16) was initialized with self-supervised 
pretrained weights and was fine-tuned with a learning rate of
$10^{-5}$, and the UNETR decoder with a learning rate of $10^{-4}$,
both with cosine annealing scheduling and a 5-epoch linear warmup.
These hyperparameters follow those reported in
ProstNFound+ \citep{wilson2026prostnfound+}.

\subsubsection{Data Augmentation}

Augmentation was applied to the training set only. Two
transformations were applied sequentially: random affine
translation with a maximum displacement of 20\% in each spatial dimension, and random resized cropping to $512 \times 512$ pixels
with scale drawn uniformly from $(0.3, 1.0)$, applied with
probability $0.5$. The needle mask and prostate mask were
transformed consistently with the image to preserve spatial
correspondence. Images were normalized to $[0,1]$ by min-max
scaling and then standardized with mean $(0,0,0)$ and standard
deviation $(1,1,1)$.

\subsubsection{Class and Mask Decoder}

Following the UNETR encoder-decoder, the spatial feature
representations are passed to a two-way transformer module
\citep{kirillov2023segment} that performs bidirectional cross-attention
between image embeddings and learnable query tokens. This
produces contextualized representations that are then routed
to two task-specific decoders.

\subsubsection{Mask Decoder}

The mask decoder receives the two-way transformer output and
produces a dense spatial heatmap $P \in [0,1]^{H \times W}$
over the image. It consists of a sequence of transposed
convolutional upsampling blocks that progressively increase
spatial resolution from the compressed transformer output back
to the input image dimensions. The heatmap is activated with
sigmoid to produce per-pixel cancer probability estimates.
Core-level cancer scores are derived by averaging heatmap
activations within the needle trace region.

\subsubsection{Class Decoder}

The class decoder receives the same two-way transformer output
and produces an image-level classification score for clinically
significant PCa. A learnable classification token attends over
the image embeddings via cross-attention, and the resulting
token representation is projected through a small MLP to produce
a scalar logit $z_{\mathrm{cls}} \in \mathbb{R}$, which is
activated with sigmoid to give the predicted csPCa probability
$\hat{y}_{\mathrm{cls}} \in [0,1]$.

The two decoders are trained jointly via the multi-task loss
described in Section \ref{sec:training_objective}, with the mask
decoder supervised by core-level involvement labels and the
class decoder supervised by binary csPCa labels. At inference,
the class decoder score is used for patient-level and
image-level evaluation, while the mask decoder heatmap is
used for core-level evaluation via needle trace averaging.
During TTA, both decoders remain frozen.

\subsubsection{Evaluation Metrics}

Performance was evaluated using AUC of the receiver operating
characteristic curve at three levels: patient-level csPCa, core-level csPCa (GG $\geq$ 3
vs.\ all other cores), and core-level PCa (any cancer vs.\
benign). Core-level scores are derived by averaging heatmap
activations along the biopsy needle trace mask. For patient-level evaluation, the patient-level 
prediction is the mean of the model's core-level scores across all cores for that patient. 
Patient-level csPCa AUC reflects the classification of patients designated as csPCa (GG $\geq$ 3)
versus non-csPCa based on pathological grading.

\subsection{Test-Time Adaptation}

\subsubsection{Segmentation Head Architecture}

The segmentation head attached to the encoder for TTA consists
of two $3\times3$ convolutional layers each followed by
GroupNorm (32 groups) and ReLU activation, and a final
$1\times1$ projection to a single output channel:

\begin{equation}
  p_\psi: \mathbb{R}^{C \times H' \times W'} \rightarrow
  \mathbb{R}^{1 \times H' \times W'}
\end{equation}

where $C = 1024$ is the encoder feature dimension and the
intermediate hidden dimension is $256$. The head is randomly
initialized at the start of the test set and is never reset
throughout inference.

\subsubsection{TTA Optimizer}

The encoder adaptation and segmentation head use separate Adam
optimizers. The encoder optimizer uses a learning rate of
$\eta_{\mathrm{enc}} = \eta_{\mathrm{TTT}} \times 0.1$ and the
segmentation head optimizer uses $\eta_{\mathrm{head}} =
\eta_{\mathrm{TTT}}$. Both optimizers are stateful and persist
across all test samples, accumulating first and second moment
estimates throughout the test set. Test samples follow a fixed sequential order by patient ID. 
For the sensitivity analysis results reported in (Table \ref{tab:ordering}), the following random seeds
were used: 10, 20, 40, 60, and 70.

\subsubsection{TTA Loss}

The TTA objective is a combined Dice and binary cross-entropy
loss applied between the predicted segmentation $\hat{s}_j$ and
the pseudo-mask $\hat{m}_j$:

 \begin{equation}
   \mathcal{L}_{\mathrm{TTA}} = \mathcal{L}_{\mathrm{Dice}}(\hat{s}_j, \hat{m}_j)
   + \lambda \mathcal{L}_{\mathrm{BCE}}(\hat{s}_j, \hat{m}_j),
   \label{ap_eq:loss}
 \end{equation}

Segmentation quality during TTA is monitored using the Dice
similarity coefficient between binarized predictions
($\hat{s}_j > 0.5$) and binarized pseudo-masks ($\hat{m}_j > 0.5$),
computed using the MONAI \texttt{DiceMetric} \citep{cardoso2022monai} with background
included. In our setting, we fix $\lambda =1$.

\subsubsection{Block Selection}

The encoder blocks are partitioned into three equal groups
(first third, middle third, last third) based on total block
count $L$:

\begin{equation}
  \text{first} = \text{blocks}[0 : L/3], \quad
  \text{middle} = \text{blocks}[L/3 : 2L/3], \quad
  \text{last} = \text{blocks}[2L/3 : L]
\end{equation}

For the DINOv3 ViT-L/16 encoder ($L=24$), the first third
comprises blocks $0-7$. The optimal number of blocks to adapt
was $n=6$, corresponding to the first 6 blocks (25\% of the
encoder). For the MedSAM encoder in ProstNFound+ ($L=12$),
the equivalent proportion yields $n=3$ blocks.
When restricting to
normalization parameters only, only the affine parameters
($\gamma$, $\beta$) of LayerNorm, BatchNorm, and GroupNorm
layers are included. When restricting to bias terms only,
only the bias parameters of Linear and Conv2d layers are
included.

\begin{table}[t]
  \centering
  \caption{Selected TTA hyperparameters.}
  \begin{tabular}{lc}
    \toprule
    \textbf{Hyperparameter} & \textbf{Value} \\
    \midrule
    TTA learning rate ($\eta_{\mathrm{TTT}}$) & $10^{-4}$ \\
    Encoder learning rate ($\eta_{\mathrm{enc}}$) & $10^{-5}$ \\
    Inner steps ($\tau$) & 10 \\
    Blocks to adapt ($n$, DINOv3) & 6 \\
    Blocks to adapt ($n$, MedSAM) & 3 \\
    Segmentation head hidden dim & 256 \\
    \bottomrule
  \end{tabular}
  \label{tab:hparams}
\end{table}

\subsubsection{Hyperparameter Selection}

TTA hyperparameters were selected on a held-out split of the
source domain training data ($80\%$ train / $20\%$ validation,
patient-level split across 693 patients). The selection
criterion was core-level PCa AUC on the validation split.
The selected hyperparameters were fixed for all LOCO evaluation
folds without center-specific tuning. Optimal values are
summarized in Table \ref{tab:hparams}.

\subsection{Real-Time Inference Latency}
Table \ref{tab:latency} reports wall-clock inference time per image for 
all methods on a single NVIDIA L40S GPU (48 GB). ANT requires 
$0.90 \pm 0.06$~seconds per image ($\tau = 10$ inner optimisation steps), 
compared to $0.07 \pm 0.01$~seconds for standard inference without 
adaptation. While ANT is slower than all baseline methods, its sub-second 
inference time is compatible with the procedural cadence of 
micro-ultrasound guided biopsy, where needle positioning and firing 
inherently require several seconds per core.

\begin{table}[!t]
\centering
\caption{Mean wall-clock inference time per image (seconds $\pm$ std) 
on a single NVIDIA L40S GPU (48~GB). ANT includes $\tau = 10$ inner 
optimisation steps per image.}
\label{tab:latency}
\footnotesize
\begin{tabular}{lc}
\toprule
\textbf{Method} & \textbf{Time (s)} \\
\midrule
No adaptation   & $0.07 \pm 0.01$ \\
TENT            & $0.11 \pm 0.03$ \\
EATA            & $0.13 \pm 0.01$ \\
ROID            & $0.19 \pm 0.03$ \\
SAR             & $0.22 \pm 0.02$ \\
CoTTA           & $0.26 \pm 0.02$ \\
MEMO            & $0.29 \pm 0.02$ \\
\midrule
\textbf{ANT (ours)} & $\mathbf{0.90 \pm 0.06}$ \\
\bottomrule
\end{tabular}
\vspace{0.5em}
\end{table}

\subsection{Long-Term Sequential Stability}

To assess ordering sensitivity, ANT was evaluated under five independent 
random permutations of the test set and compared against five runs under 
the fixed original ordering, reporting mean AUC and standard deviation 
across seeds (Table \ref{tab:ordering}). Performance is stable across 
orderings, with differences of less than 2 AUC points across all metrics 
and centres, and overlapping standard deviations in all cases, 
demonstrating that ANT is robust to the order in which test images are 
presented during episodic adaptation.

\begin{table}[!t]
\centering
\caption{Ordering sensitivity analysis. Mean AUC (\%) $\pm$ std across 
five random seeds for fixed and randomly permuted test set orderings.}
\label{tab:ordering}
\resizebox{\textwidth}{!}{%
\begin{tabular}{lcccccc}
\toprule

& \multicolumn{2}{c}{\textbf{Patient-Level csPCa AUC (\%)}}
& \multicolumn{2}{c}{\textbf{Core-Level csPCa AUC (\%)}}
& \multicolumn{2}{c}{\textbf{Core-Level PCa AUC (\%)}} \\

\cmidrule(lr){2-3} \cmidrule(lr){4-5} \cmidrule(lr){6-7}

{\textbf{Ordering}}

& \textbf{B} & \textbf{C}
& \textbf{B} & \textbf{C}
& \textbf{B} & \textbf{C} \\
\midrule
Fixed order
    & $79.9 \pm 0.03$ & $84.1 \pm 0.08$
    & $81.1 \pm 0.02$ & $88.0 \pm 0.06$
    & $73.1 \pm 0.03$ & $72.3 \pm 0.03$ \\
Random order
    & $80.0 \pm 0.04$ & $82.3 \pm 0.03$
    & $80.6 \pm 0.04$ & $85.8 \pm 0.08$
    & $72.9 \pm 0.01$ & $72.0 \pm 0.05$ \\
\bottomrule
\end{tabular}%
}
\vspace{0.5em}
\end{table}

\subsection{Calibration Analysis}

\begin{table}[!t]
\centering
\caption{Calibration analysis on Center B ($n=1{,}118$ cores). Expected 
Calibration Error (ECE) is reported for the unadapted baseline (PNF+) and 
ANT. $\Delta$ECE = ANT ECE $-$ Baseline ECE; negative values indicate 
improved calibration.}
\label{tab:calibration}
\begin{tabular}{lcccc}
\toprule
\textbf{Subgroup (Involvement)} & \textbf{$n$} & \textbf{Baseline ECE (\%)} & \textbf{ANT ECE (\%)} & \textbf{$\Delta$ECE (\%)} \\
\midrule
Overall                     & 1{,}118 & 7.6  & 2.7  & $-4.9$ \\
\midrule
Low     & 161     & 29.7 & 32.0 & $+2.2$ \\
Mid     & 93      & 19.6 & 16.5 & $-3.0$ \\
High    & 131     & 14.3 & 13.8 & $-0.5$ \\
\bottomrule
\end{tabular}
\end{table}

We assessed the effect of ANT on model calibration using Expected Calibration Error (ECE) on Center B, 
which provides the most representative subgroup distribution. ANT substantially improves overall calibration 
relative to the PNF+ baseline (ECE: $2.7\%$ vs.\ $7.6\%$, $\Delta\text{ECE}=-4.9\%$). For cancer-positive 
cores stratified by involvement tertile, ANT maintains or improves calibration in the mid ($\Delta\text{ECE}=-3.0\%$) 
and high ($\Delta\text{ECE}=-0.5\%$) involvement subgroups. A modest increase in ECE is observed in the low involvement subgroup ($\Delta\text{ECE}=+2.2\%$), suggesting slight overconfidence on cores with minimal cancer extent, which is consistent with the inherent difficulty of detecting low-involvement cancer under domain shift.

 \end{document}